\documentclass[sigconf]{acmart}

\usepackage{times}
\usepackage{colortbl}
\usepackage{amsmath}
\usepackage{mathtools}
\usepackage{amsthm}

\usepackage{algorithm}
\usepackage{algorithmic}
\usepackage{microtype}
\usepackage{graphicx}
\usepackage{subfigure}
\usepackage{booktabs} 
\usepackage{multirow}
\usepackage{array}
\usepackage{colortbl}
\usepackage{xcolor}
\usepackage{array}
\usepackage{geometry}
\usepackage{makecell}

\usepackage{amssymb}
\usepackage{pifont}
\usepackage{subcaption} 
\usepackage{longtable}

\usepackage[T1]{fontenc}

\usepackage[utf8]{inputenc}
\usepackage{pifont}

\usepackage{microtype}
\usepackage{makecell}

\usepackage{graphicx}
\AtBeginDocument{%
  }

\setcopyright{acmlicensed}
\copyrightyear{2025}
\acmYear{2025}
\acmDOI{XXXXXXX.XXXXXXX}
\acmConference[]{A Versatile Multimodal Agent for Multimedia Content Generation}
\acmBooktitle{}

\acmSubmissionID{4}

\begin{document}

\title{A Versatile Multimodal Agent for Multimedia Content Generation}

\author{Daoan Zhang}
\email{daoan.zhang@rochester.edu}
\affiliation{%
  \institution{University of Rochester}
  \country{USA}
}

\author{Wenlin Yao}
\email{wenlinyao@global.tencent.com}
\affiliation{%
  \institution{Tencent AI Lab, Bellevue}
  \country{USA}
}

\author{Xiaoyang Wang}
\email{shawnxywang@global.tencent.com}
\affiliation{%
  \institution{Tencent AI Lab, Bellevue}
  \country{USA}
}

\author{Yebowen Hu}
\email{yebowen.hu@ucf.edu}
\affiliation{%
  \institution{University of Central Florida}
  \country{USA}
}

\author{Jiebo Luo}
\email{jluo@rochester.edu}
\affiliation{%
  \institution{University of Rochester}
  \country{USA}
}

\author{Dong Yu}
\email{dyu@global.tencent.com}
\affiliation{%
  \institution{Tencent AI Lab, Bellevue}
  \country{USA}
}

\renewcommand{\shortauthors}{Daoan et al.}

\begin{abstract}
With the advancement of AIGC (AI-generated content) technologies, an increasing number of generative models are revolutionizing fields such as video editing, music generation, and even film production. However, due to the limitations of current AIGC models, most models can only serve as individual components within specific application scenarios and are not capable of completing tasks end-to-end in real-world applications. In real-world applications, editing experts often work with a wide variety of images and video inputs, producing multimodal outputs---a video typically includes audio, text, and other elements. This level of integration across multiple modalities is something current models are unable to achieve effectively. However, the rise of agent-based systems has made it possible to use AI tools to tackle complex content generation tasks.
To deal with the complex scenarios, in this paper, we propose a \textbf{MultiMedia-Agent} designed to automate complex content creation. Our agent system includes a data generation pipeline, a tool library for content creation, and a set of metrics for evaluating preference alignment. Notably, we introduce the skill acquisition theory to model the training data curation and agent training. We designed a two-stage correlation strategy for plan optimization, including self-correlation and model preference correlation. 
Additionally, we utilized the generated plans to train the MultiMedia-Agent via a three stage approach including base/success plan finetune and preference optimization. The comparison results demonstrate that the our approaches are effective and the MultiMedia-Agent can generate better multimedia content compared to novel models.
\end{abstract}




\maketitle

\section{Introduction}
\label{sec:intro}

AI-generated content—including images, videos, audio, and more—has increasingly permeated various aspects of everyday life~\citep{liu2024evalcrafter, esser2024scaling, tang2023video}. However, real-world demands are both complex and diverse. For example, in the field of video generation, a user's input may extend beyond text or a single image to include additional materials such as images, music, etc., requiring the model to integrate these elements into a coherent video. Likewise, the desired output might not only consist of video frames but also incorporate background music, voiceovers, subtitles, and other elements. One promising solution is to employ an agent system that comprehends user needs while seamlessly integrating various downstream tools to manage complex inputs and outputs~\citep{wang2024modaverse, wang2024mllm}. Although some existing agent systems~\citep{shen2024hugginggpt} can invoke multiple tools for content generation, they are not specifically designed for this purpose and do not fully address the diverse requirements of real-world applications.

In this paper, we explore whether a multimodal agent can learn to handle such complex multimedia content creation workflows in a manner akin to human learning. Specifically, we investigate whether the multimodal agent can progressively acquire complex skills from scratch by following the stages of \textbf{Skill Acquisition Theory}~\citep{dekeyser2020skill}, which mirrors the human step-by-step learning process. According to this theory, skill acquisition occurs in three stages: \begin{enumerate} \item \textbf{Cognitive Stage}: Beginners learn the fundamental operations and core concepts. \item \textbf{Associative Stage}: Learners engage in targeted practice, refining their skills through repeated operations. \item \textbf{Autonomous Stage}: Continuous feedback—through both self-correction and external guidance—leads to proficient, automatic performance. \end{enumerate}

To investigate this process, we first developed a multimedia content playground grounded in real-world scenarios and equipped with a feedback mechanism. This playground encompasses \textbf{18 common multimodal generation scenarios} and includes a \textbf{content creation tool library} that supports the editing, generation, and retrieval of images, videos, audio, speech, and text. To train our multimodal agent, we generated hierarchical plans corresponding to the three stages of Skill Acquisition Theory. Initially, we employed GPT-4o as a teacher to generate a base plan for each task. Recognizing that these base plans might not always execute successfully, we refined them further in two steps. First, GPT-4o performed self-reflection and self-correction to enhance the quality of the base plans. Then, external preference models were introduced to evaluate and further optimize these plans. This iterative process yielded three distinct levels of plans for training the multimodal agent.

Guided by Skill Acquisition Theory, our training process mirrors the human learning trajectory. In the \textbf{Model Cognitive Stage}, we fine-tuned the agent using all generated plans, enabling it to quickly grasp the functions, output formats, and basic operational principles of various tools—much like a beginner acquiring foundational knowledge. In the \textbf{Model Associative Stage}, we fine-tuned the agent using only the successfully executed plans, allowing it to learn advanced logic such as workflow composition and tool interrelationships. Finally, in the \textbf{Model Autonomous Stage}, post-training was conducted using paired preference data derived from the model's evaluations. This final stage empowers the agent not only to complete tasks effectively but also to integrate aesthetic and emotional considerations—such as human preferences—into its tool execution.

It is important to highlight that our MultiMedia-Agent significantly differs from previous models and methods. As summarized in Table~\ref{tab:my_label}, while many existing systems support multimodal content generation, their approaches vary. For instance, HuggingGPT~\citep{shen2024hugginggpt} handles multimodal understanding indirectly via API calls, whereas NExT-GPT~\citep{wu2023next} and MLLM-Tool directly interpret multimodal data. AutoDirector~\citep{ni2024autodirector} is limited to text input and cannot generate content based on user-provided materials—unlike our model, which accepts multiple and multimodal inputs. In terms of planning, ToolLLM~\citep{qin2023toolllm} and HuggingGPT can orchestrate the use of multiple tools based on user instructions, while NExT-GPT and ModaVerse~\citep{wang2024modaverse} produce only a single content piece per forward pass. Regarding multimodal interaction—the ability to generate and integrate various modalities based on user needs—current systems like AutoDirector (limited to video scenarios) fall short of our model’s capabilities. Moreover, none of the existing models address preference alignment~\citep{zhang2024seppo} in content generation, whereas MultiMedia-Agent incorporates human-preference-based evaluation models to manage this aspect.

In summary, our contributions are as follows: \begin{enumerate} \item We develop a plan generation system for multimedia content creation, including a data generation pipeline, a comprehensive tool library, and tailored evaluation metrics. \item We introduce a two-stage plan curation strategy for multimedia content generation, utilizing self-reflection and preference model-based optimization. \item Based on Skill Acquisition Theory, We propose \textbf{Agent Skill Acquisition}, a three-stage training pipeline for a multimedia content generation agent that enables it to learn and generate complex plans from scratch. \end{enumerate}

\begin{table*}[t!]
    \centering
    \begin{tabular}{c|c|c|c|c|c}
        \hline
         & \makecell{Multimodal \\ Generation} & \makecell{Multimodal \\ Understanding} & \makecell{Planning \\ Ability} & \makecell{Multimodal \\ Interaction} & \makecell{Preference \\ Alignment}
                \\
        \hline
        ToolLLM & \ding{55} & \ding{55}  & \ding{51} & \ding{55} & \ding{55} \\
        \hline
        HuggingGPT & \ding{51} & \ding{55}  & \ding{51} & \ding{55} & \ding{55} \\
        \hline
        NExT-GPT & \ding{51} & \ding{51}  & \ding{55} & \ding{55} & \ding{55} \\
        \hline
        ModaVerse & \ding{51} & \ding{51}  & \ding{55} & \ding{55} & \ding{55} \\
        \hline
        AutoDirector & \ding{51} & \ding{55}  & \ding{51} & \ding{51} & \ding{55} \\
        \hline
        MultiMedia-Agent & \ding{51} & \ding{51}  & \ding{51} & \ding{51} & \ding{51} \\
        \hline
    \end{tabular}
    \caption{A comparison of our MultiMedia-Agent with notable tool agents or content creation agents.}
    \label{tab:my_label}
\end{table*}

\section{Related Work} 

\subsection{Tool Agent}
With the rise of large language models (LLM) agents~\citep{mei2024llm, liu2023dynamic, liu2024agentlite, zhao2024expel}, enabling agents to call external APIs to solve user problems has become a crucial research topic. Toolformer~\citep{schick2024toolformer} pioneered the exploration of connecting LLMs~\cite{hu2024define} with external tools. HuggingGPT~\citep{shen2024hugginggpt} leveraged an agent to call HuggingFace’s API, allowing it to solve a wide range of complex problems. Subsequent research has extended this integration to fields like healthcare support~\citep{ma2023understanding}, code synthesis~\citep{wang2024executable}, and web searching~\citep{ma2023laser}. ToolLLM~\citep{qin2023toolllm} focused on executing complex tasks in real-world scenarios. GPT4Tools~\citep{yang2024gpt4tools} and Visual ChatGPT~\citep{wu2023visual} integrated visual foundation models after decomposing tasks into manageable components. For multimodal tool agents, MLLM-Tool~\citep{wang2024mllm} employed multimodal large models as agents to call Hugging Face APIs. Similarly, ModaVerse~\citep{wang2024modaverse} used multimodal large models for any-to-any generation. In our MultiMedia-Agent, we focus primarily on the planning and alignment capabilities of multimodal agents, aiming to enhance content generation quality.

\subsection{Any-to-any generation}

The earliest any-to-any model was CoDi~\citep{tang2024any}, followed by NextGPT and EMU~\citep{sun2023emu}, which introduced further improvements in data and model design. EMU2~\citep{sun2024generative} introduced a unified autoregressive objective to predict the next multimodal element, either by regressing visual embeddings or classifying textual tokens. CM3Leon~\citep{yu2023scaling} and Chameleon~\citep{team2024chameleon} used mixed image and text data to train token-based autoregressive models. More recently, TransFusion~\citep{zhou2024transfusion} and Show-o~\citep{xie2024show} combined large language models with diffusion models to enhance performance.

However, any-to-any models are typically limited to generating a single modality without considering the relationships and connections between modalities. This is precisely the area that our MultiMedia-Agent focuses on, emphasizing the interplay between different modalities for richer content generation.

\section{Data Curation}

In this section, we systematically analyze the procedures used to construct our dataset for generating complex multimodal content. First, we built a multimodal tool library from which the agent can select appropriate tools to form a plan. Next, we constructed differentiated plans to address diverse requests and feedback. Finally, we designed a series of metrics to evaluate the quality of content generated by these plans, providing preference feedback that enables both assessment and ranking.

\subsection{Multi-media Tasks}

Because no existing dataset fully captures real-world multimodal demand-solution scenarios, we constructed 18 different tasks based on various practical needs, as shown in Table~\ref{tasks}. For instance, a user might wish to automatically convert a series of photos into a video slideshow for a wedding or event montage (multi-images-to-video). Another scenario could involve combining a set of photos with chosen background music to produce a travel memory video (image-audio-to-video). In total, we designed 18 task types involving image, video, audio, speech, and text.

To create our dataset, we first gathered publicly available multimedia data, then used GPT-4o to generate user queries under varying conditions by integrating this data with the relevant task type information. Finally, we obtained a diverse dataset of multimedia tasks, each mapped to specific user queries and corresponding multimedia data.

\begin{table*}[t!]
    \centering
    \begin{tabular}{c|c|c}
\hline
Audio/Video to Audio & Audio/Video to Text & Audio/Video to Video \\
\hline
Image/Audio to Text & Image/Audio to Video & Image/Video to Audio \\
\hline
Image/Video to Text & Image/Video to Video & Multiple Audios to Image \\
\hline
Multiple Audios to Text & Multiple Audios to Video & Multiple Images to Audio \\
\hline
Multiple Images to Text & Multiple Images to Video & Multiple Videos to Audio \\
\hline
Multiple Videos to Image & Multiple Videos to Text & Multiple Videos to Video \\
\hline
\end{tabular}
\caption{18 real world task types.}
\label{tasks}
\end{table*}

\subsection{Tool Library}

Given the complex interconnections among different modalities, we built our tool library from three main perspectives: Multimodal Understanding Tools, Generative/Editing Tools, and Auxiliary Tools. A detailed overview can be found in supplementary material.

\textbf{Multimodal Understanding Tools.}
An effective agent should be able to perceive its environment before acting. Hence, we designed understanding models for each modality, allowing the agent to analyze data of various types and facilitating more effective plan curation. Specifically, we introduced five any-to-text models—one for each modality (image, video, speech, audio, and text).

\textbf{Generative/Editing Tools.}
Our agent also needs to generate and edit data across different modalities. To this end, we provided a suite of generation and editing tools for images, video, audio, and speech. We additionally integrated several non-deep-learning tools—such as video transition effects and audio effects—to ensure comprehensive multimedia editing capabilities.

\textbf{Auxiliary Tools.}
We included essential auxiliary tools for common multimedia operations, such as video-to-video concatenation, video-to-audio synchronization, video retrieval, and other fundamental tasks.

\begin{figure*}
    \centering
    \includegraphics[width=1.0\linewidth]{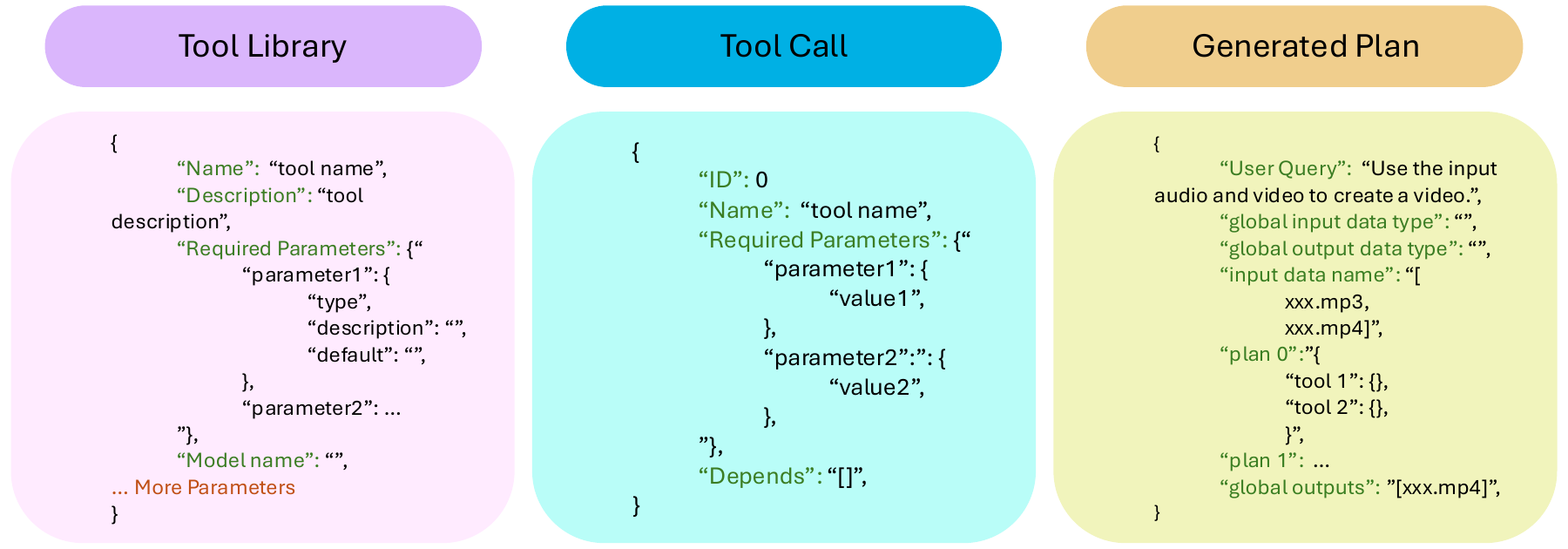}
    \caption{Formats for tool library and generated plan.}
    \label{fig:plan}
\end{figure*}

We organized the information for each tool into JSON file. The keys in the prompt consist of the following part: \textbf{Tool name and execution model name:} We first defined the tool names and their corresponding models. When designing the tool names, we considered that our agent involves multiple modalities and various input-output models, which can easily lead to incorrect file formats being generated during the planning stage. To address this, we fixed the file formats for the four modalities as follows: Image: \textit{.png}; Video: \textit{.mp4}; Audio $\&$ Speech: \textit{.mp3}; Text: \textit{.txt}. We also included both the input and output formats in the tool name, for example, \textit{text\_txt\_to\_video\_mp4}, to ensure more stable plan curation. Additionally, the JSON file defines the model names associated with each tool, which are used to index the models during execution. \textbf{Required parameters:} Due to the complexity of our tasks and data, for each required parameter, we provide a detailed description of its purpose. For example, in the object removal tool, where the input parameters include a text description and the input image name, the required input parameters are defined as: \textit{\{"text", "description of the object to be removed"\}} and \textit{\{"image", "image file from which object needs to be removed"\}}. This approach helps ensure that the model can output the tool information more accurately. \textbf{Tool Description:} Including the functionality and description of the tool is essential to better prompt the agent model. We show the format of the tool library; tool calling function and Generated plan in the Figure. \ref{fig:plan}.

\subsection{Hierarchical Plan Curation}

Once we have constructed the user queries, corresponding multimedia data, and the tool library, we utilize GPT-4o to generate the plans. These plans are organized into a list of dictionaries, where each dictionary contains information about a specific tool.

To enable the multimodal agent to better learn complex skills, we need to incorporate feedback into the model's training data. Unlike conventional tool agents, content-generation-oriented tool agents must not only consider the execution success rate of the plan but also ensure that the generated multimedia content meets human needs and aesthetic standards. In other words, the agent we are training must not only possess the ability to complete complex tasks using tools but also be responsible for the outcomes produced by these tools, ensuring that the results align with human needs and preferences.

To address this, we designed a two-stage correlation approach for tool plan curation. After generating the base plan, we first employ GPT-4o to perform self-correction, identifying issues within the plan and optimizing it to obtain a self-corrected plan. Next, we execute the self-corrected plan to produce the multimedia result. We then apply a series of model-based preference evaluation metrics to assess the quality of the multimedia output. Based on these metrics, the LMM further refines the plan to optimize it, ultimately yielding the final plan.

\subsubsection{Two-stage correlation of plan curation}

\begin{figure*}
    \centering
    \includegraphics[width=\linewidth]{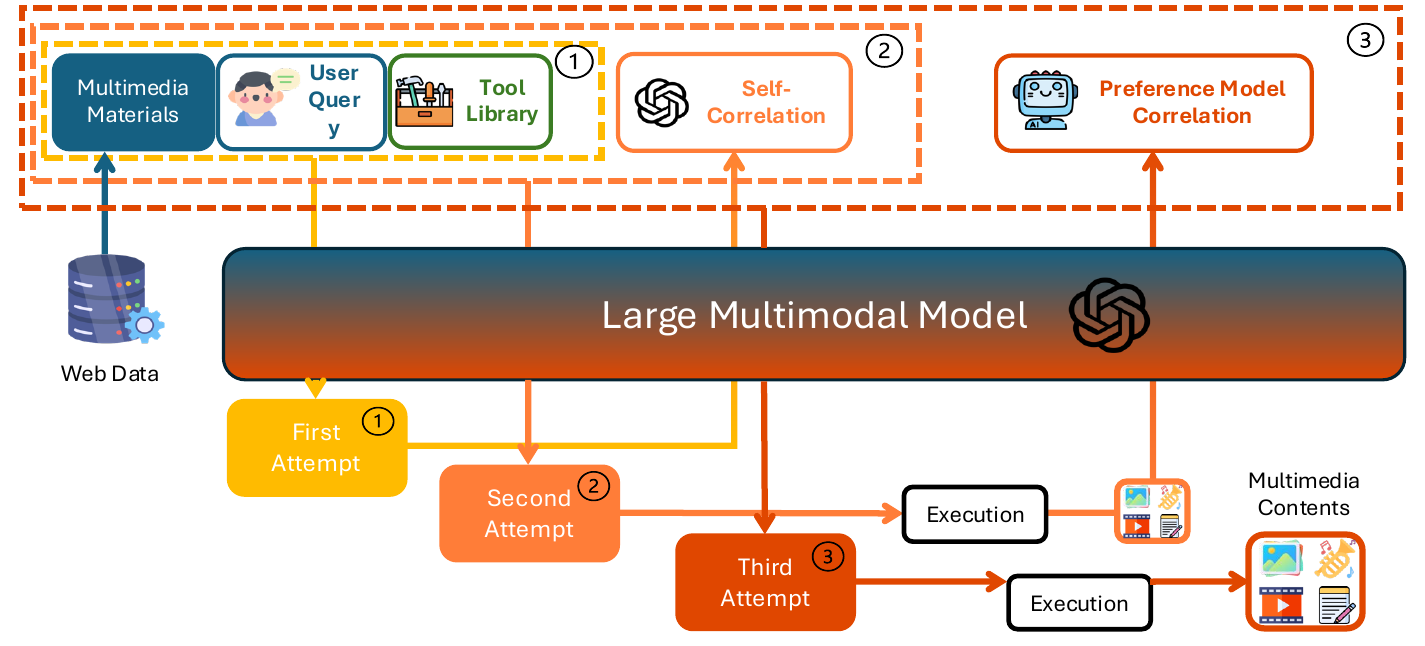}
    \caption{Two-stage correlation of plan curation for content creation.}
    \label{fig:enter-label}
\end{figure*}

\textbf{Stage 1: Self-correlation}

After inputting the user query, materials, and tool information into GPT-4o to generate the base plan, we further prompt GPT-4o to evaluate the quality of the current plan based on all available information. Our evaluation criteria focus on two main aspects: \textbf{Plan execution success rate:} We prompt GPT-4o to assess whether the plan can be successfully executed, and if not, make necessary modifications. \textbf{Inclusion of user-requirement-aligned, common-sense optimizations:} We assess whether the plan includes additional optimization tools that meet the user's needs and adhere to common sense, such as adding background music to a video or incorporating voiceovers in audio generation. This ensures the plan is both executable and aligned with user expectations.

\textbf{Stage 2: Model Preference Correlation}

To further evaluate the generated plans, in Stage 2 we assess the results of content generation plans using model-based preference feedback metrics for the four output modalities: image, video, audio, and text. Our evaluation focuses on three primary aspects: whether the generated multimedia content meets human needs, conveys emotional expression, and aligns across modalities.

\begin{itemize}
    \item \textbf{Text output metrics.} We use GPT-4o to evaluate the alignment between the input and output content.
    \item  \textbf{Image output metrics.} GPT-4o assesses whether the images meet human needs and convey emotions, while Pick Score~\cite{kirstain2023pick} is used to evaluate aesthetics.
    \item \textbf{Audio output metrics.} We apply speech-to-text and audio-to-text models to convert the audio into text, and GPT-4o evaluates the fulfillment of human needs and emotional expression based on user requirements and input content.
    \item \textbf{Video output metrics.} Similar to image outputs, GPT-4o evaluates whether the video meets human needs and conveys emotions, while Dover Score~\citep{wu2023dover} is used for aesthetic and quality evaluation. If the video includes embedded audio, we apply the same evaluation methods used for audio outputs. Additionally, we introduce a audio-video alignment metric, where GPT-4 scores the alignment between the transcribed audio text and the video content.
\end{itemize}

By integrating these metrics, we provide a comprehensive evaluation of any type of plan execution output, thereby reflecting the overall quality of the plan. We use the optimized plans from Stage 1 to generate multimedia content, then apply the above metrics for evaluation. The evaluation results are fed back to GPT-4o, which, based on this feedback and prior information, generates a new plan. A sample generated plan can be found in supplementary material.
Based on our plan-generation approach, for each case we created three sets of plans, corresponding to Plan 1, Plan 2, and Plan 3.

\subsection{Data Statistics}

In this section, we primarily present the statistics of the dataset we constructed, including success rate, average steps, and average metrics. For each task type, we generated 1260 user requests under different conditions. For each individual user request, we constructed three plans. We first calculated the number of steps in each plan for each task type, as shown in Table \ref{tab:step}. Here, "T" represents "Text," "A" represents "Audio," "V" represents "Video," "I" represents "Image," and "M" represents "Multi-input." Among them, '-' represents generation. For example, 'AV-A' means using Audio + Video to generate Audio.
We can observe that for more complex tasks like video generation, the plans tend to include more steps to complete the task, whereas for text generation, the model requires fewer steps to accomplish the task. Additionally, as the plans are optimized, the number of steps required to complete the tasks increases, indicating that Self-correlation and Model Preference Correlation introduced more tool usage in the plan generation process.

We further illustrate the success rates of the generated plans in the supplementary materials. When combined with the number of steps in each plan, it becomes evident that as the number of steps in a plan increases, the success rate of the plan decreases. 

\begin{table*}[h!]
\centering

\begin{tabular}{c|c|c|c|c|c|c|c|c|c}
    \hline
     & \textbf{AV-A} & \textbf{AV-T} & \textbf{AV-V} & \textbf{IA-T} & \textbf{IA-V} & \textbf{IV-A} & \textbf{IV-T} & \textbf{IV-V} & \textbf{MA-I} \\
    \hline
    \textbf{Plan 1} & 5.8 & 2.9 & 4.1 & 3.0 & 4.8 & 4.3 & 3.0 & 6.3 & 5.1 \\
    \hline  
    \rowcolor{gray!25}\textbf{Plan 2} & 6.1 & 3.1 & 5.4 & 3.0 & 5.6 & 8.4 & 3.0 & 6.6 & 5.2 \\
    \hline
   \rowcolor{gray!40} \textbf{Plan 3} & 6.2 & 3.1 & 5.6 & 3.0 & 6.2 & 9.2 & 3.1 & 7.8 & 6.3 \\
    \hline

\end{tabular}

\begin{tabular}{c|c|c|c|c|c|c|c|c|c}
    \hline
    & \textbf{MA-T} & \textbf{MA-V} & \textbf{MI-A} & \textbf{MI-T} & \textbf{MI-V} & \textbf{MV-A} & \textbf{MV-I} & \textbf{MV-T} & \textbf{MV-V} \\
    \hline
    \textbf{Plan 1} & 4.0 & 8.1 & 4.2 & 4.1 & 8.5 & 7.6 & 12.0 & 4.1 & 6.0 \\
    \hline
    \rowcolor{gray!25}\textbf{Plan 2} & 4.1 & 9.4 & 5.5 & 4.1  & 8.8 & 8.0 & 12.2 & 4.1 & 6.4 \\
    \hline
    \rowcolor{gray!40}\textbf{Plan 3} & 4.4 & 11.8 & 8.2 & 4.5 & 10.6 & 9.2 & 12.8 & 4.1 & 7.4 \\
    \hline
\end{tabular}
\caption{Average steps for different tasks in the training set.}
\label{tab:step}
\end{table*}

\section{MultiMedia-Agent}

\subsection{Agent Skill Acquisition}

We further used our data to train an multimodal agent. To better encode the tool information and the behaviors from the plan into the multimodal model, we designed a  three-stage training approach based on skill acquisition theory.

\begin{enumerate}
    \item \textbf{Model Cognitive Stage.} At this stage, the agent primarily focuses on learning the basic usage of tools and understanding the input-output JSON formats. We trained the model using all available data.

    \item \textbf{Model Associative Stage.} At this stage, we trained the model using only successful plans, the agent begins to learn established action trajectories from successfully executed plans to ensure smooth execution and accurate output of future plans. 
    
    \item \textbf{Model Autonomous Stage.} At this stage, the agent not only needs to develop the ability to synthesize complex plans but also must ensure that the generated content aligns with human aesthetics and preferences. So, we categorized the plans into winning and losing plans based on the metric model's scores. Then, we applied DPO (Direct Preference Optimization) to align the model with these preferences.
\end{enumerate}

\subsection{Experiment Settings}

\begin{figure*}[t]
    \centering
    \includegraphics[width=0.9\linewidth]{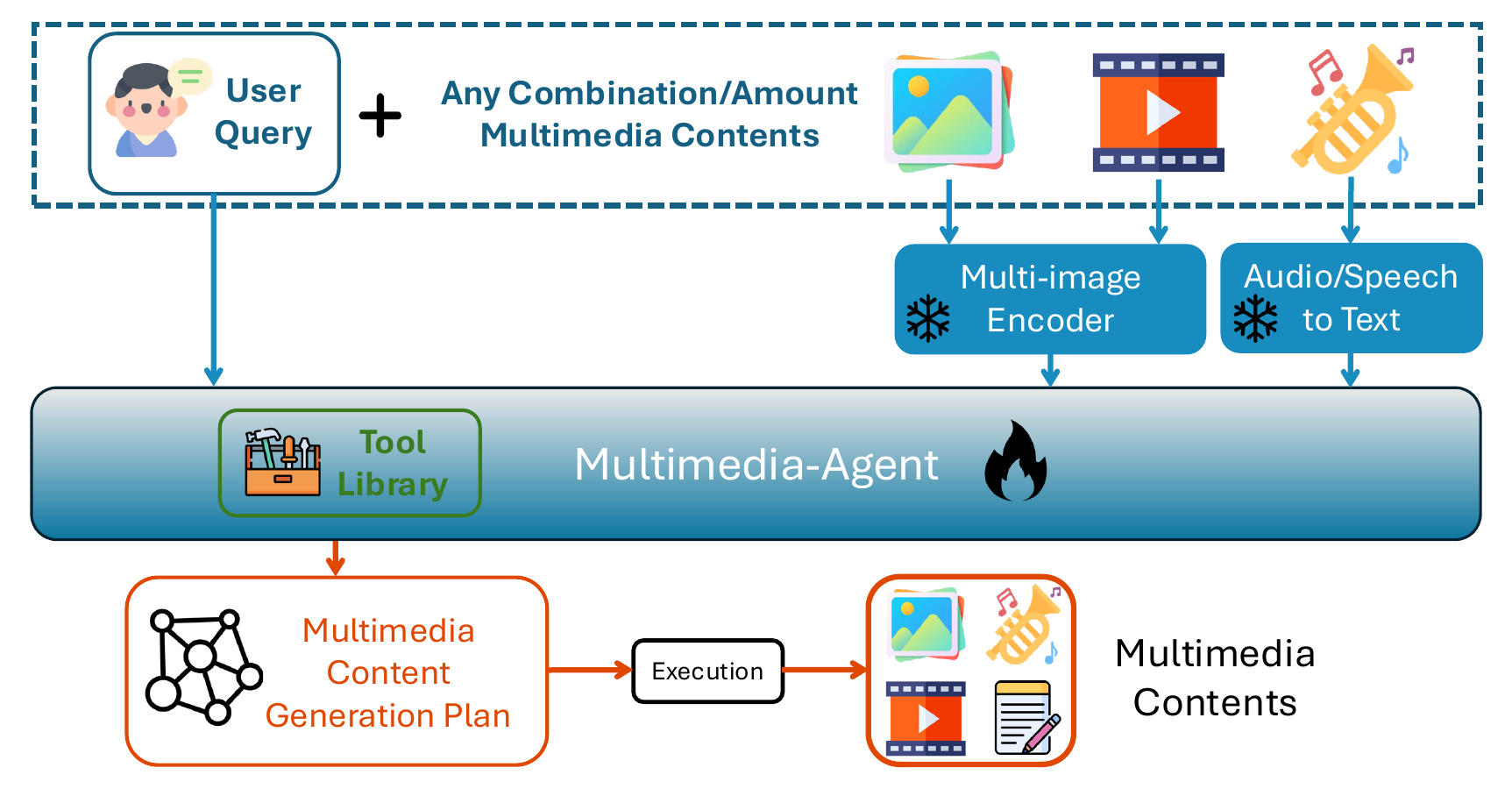}
    \caption{The detailed structure of MultiMedia-Agent.}
    \label{fig:model}

\end{figure*}

We use Minicpm-v2~\citep{yao2024minicpm} as our backbone to train out MultiMedia-Agent. The agent structure is shown in Figure \ref{fig:model}. Specifically, when processing videos, we extract 3 evenly spaced frames to represent the entire video. Since the number of input videos and images is not fixed, we concatenate all the images into a single large image before feeding it into the model. For audio and speech, we first use audio-to-text or speech-to-text models to convert the input into text, which is then passed into the LLM for further processing. The training details are attached in supplementary material.. For model validation, we generated an validation set including 180 test cases and used Claude 3.5 Sonnet, GPT-4o, GPT-4omini as a comparison method. The metrics we selected for evaluation included not only success rate but also model preference feedback. For evaluation, we provide two variaties of metrics: \textbf{plan execution success rate and preference metrics}.

\begin{table*}[t]
    \centering
    \begin{tabular}{c|c|c|c|c|c|c|c|c|c}
        \hline
         & \textbf{AV-A} & \textbf{AV-T} & \textbf{AV-V} & \textbf{IA-T} & \textbf{IA-V} & \textbf{IV-A} & \textbf{IV-T} & \textbf{IV-V} & \textbf{MA-I} \\
        \hline
        \textbf{Claude 3.5 Sonnet} & 100 & 95 & 100 & 100 & 100 & 100 & 95 & 100 & 95 \\
         \hline  
      \textbf{GPT-4o} & 100 & 100 & 100 & 100 & 100 & 100 & 100 & 100 & 100 \\
       \hline  
         \textbf{GPT-4o-Mini} & 95 & 100 & 90 & 95 & 95 & 95 & 100 & 95 & 90 \\
        \hline  
        \rowcolor{gray!15}\textbf{MultiMedia-Agent-1} & 80 & 50 & 50 & 70 & 50 & 60 & 70 & 80 & 90 \\
        \hline
        \rowcolor{gray!25}\textbf{MultiMedia-Agent-2} & 100 & 90 & 80 & 100 & 90 & 90 & 90 & 100 & 100 \\
        \hline
        \rowcolor{gray!40}\textbf{MultiMedia-Agent-3} & 100 & 90 & 40 & 100 & 60 & 80 & 100 & 40 & 90\\
        \hline
    \end{tabular}
    \begin{tabular}{c|c|c|c|c|c|c|c|c|c}
        \hline
        & \textbf{MA-T} & \textbf{MA-V} & \textbf{MI-A} & \textbf{MI-T} & \textbf{MI-V} & \textbf{MV-A} & \textbf{MV-I} & \textbf{MV-T} & \textbf{MV-V} \\
        \hline
        \textbf{Claude 3.5 Sonnet} & 95 & 65 & 95 & 100 & 75 & 100 & 85 & 100 & 85 \\
         \hline  
   \textbf{GPT-4o} & 100 & 60 & 100 & 100 & 70 & 100 & 90 & 100 & 80 \\
         \hline  
    \textbf{GPT-4o-Mini} & 95 & 55 & 100 & 95 & 65 & 95 & 90 & 95 & 75 \\
        \hline
        \rowcolor{gray!15}\textbf{MultiMedia-Agent-1} & 90 & 10 & 70 & 80  & 50 & 50 & 70 & 60 & 60 \\
        \hline
        \rowcolor{gray!25}\textbf{MultiMedia-Agent-2} & 100 & 40 & 80 & 100  & 70 & 80 & 90 & 90 & 80 \\
        \hline
        \rowcolor{gray!40}\textbf{MultiMedia-Agent-3} & 70 & 30 & 50 & 90 & 40 & 90 & 80 & 90 & 80 \\
        \hline
    \end{tabular}
    \caption{Comparison of \textbf{plan execution success rate} for \textbf{all tasks}. Models are Claude 3.5 Sonnet, GPT-4o, GPT-4o-Mini, and MultiMedia-Agent.}
    \label{tab:comparison}
\end{table*}

\begin{table*}[h]
\centering
\begin{tabular}{c|c|c|c|c|c|c}
    \hline
    & \textbf{MA-T} & \textbf{MV-T} & \textbf{IV-T} & \textbf{MA-T} & \textbf{MI-T} & \textbf{MV-T} \\
    \hline
    \textbf{Claude 3.5 Sonnet} & 4.5 & 4.1 & 4.2 & 3.7 & 4.0 & 4.8 \\
    \hline
     \textbf{GPT-4o} & 4.5 & 3.8 & 4.2 & 3.7 & 4.0 & 4.7 \\
     \hline
     \textbf{GPT-4oMini} & 4.2 & 3.8 & 4.0 & 3.7 & 3.6 & 4.2 \\
    \hline
    \rowcolor{gray!15}\textbf{MultiMedia-Agent-1} & 4.4 & 3.7 & 4.0 & 3.8 & 4.1 & 4.5 \\
    \hline
    \rowcolor{gray!25}\textbf{MultiMedia-Agent-2} & 4.4 & 3.7 & 4.0 & 3.8 & 4.1 & 4.5 \\
    \hline
    \rowcolor{gray!40}\textbf{MultiMedia-Agent-3} & 4.6 & 3.9 & 4.1 & 3.9 & 4.2 & 4.6 \\
    \hline
\end{tabular}

\begin{tabular}{c|c|c|c|c|c|c}
    \hline
    & \textbf{MA-T} & \textbf{MV-T} & \textbf{IV-T} & \textbf{MA-T} & \textbf{MI-T} & \textbf{MV-T} \\
    \hline
    \textbf{Claude 3.5 Sonnet} & 3.8 & 3.9 & 3.6 & 4.0 & 4.4 & 3.9 \\
    \hline
    \textbf{GPT-4o} & 3.8 & 3.9 & 3.6 & 4.0 & 4.2 & 3.6 \\
    \hline
    \textbf{GPT-4oMini} & 3.8 & 3.7 & 3.6 & 4.1 & 4.1 & 3.5 \\
    \hline
    \rowcolor{gray!15} \textbf{MultiMedia-Agent-1} & 3.8 & 3.7 & 3.6 & 4.1 & 4.1 & 3.8 \\
    \hline
    \rowcolor{gray!25} \textbf{MultiMedia-Agent-2} & 3.7 & 3.6 & 3.6 & 4.1 & 4.2 & 3.6 \\
    \hline
    \rowcolor{gray!40} \textbf{MultiMedia-Agent-3} & 4.3 & 3.9 & 3.8 & 4.3 & 4.1 & 3.9 \\
    \hline
\end{tabular}

\begin{tabular}{c|c|c|c|c|c|c}
    \hline
    & \textbf{MA-T} & \textbf{MV-T} & \textbf{IV-T} & \textbf{MA-T} & \textbf{MI-T} & \textbf{MV-T} \\
    \hline
    \textbf{Claude 3.5 Sonnet} & 2.1 & 1.6 & 1.8 & 1.4 & 1.7 & 2.3 \\
    \hline
    \textbf{GPT-4o} & 2.1 & 1.6 & 1.7 & 1.4 & 1.7 & 2.3 \\
    \hline
    \textbf{GPT-4oMini} & 2.1 & 1.6 & 1.6 & 1.4 & 1.4 & 2.2 \\
    \hline
    \rowcolor{gray!15} \textbf{MultiMedia-Agent-1} & 2.0 & 1.6 & 1.8 & 1.3 & 1.6 & 2.2 \\
    \hline
    \rowcolor{gray!25} \textbf{MultiMedia-Agent-2} & 2.0 & 1.5 & 1.8 & 1.5 & 1.4 & 2.2 \\
    \hline
    \rowcolor{gray!40} \textbf{MultiMedia-Agent-3} & 2.0 & 1.9 & 1.8 & 1.6 & 1.8 & 2.2 \\
    \hline
\end{tabular}

\begin{tabular}{c|c|c|c|c|c|c}
    \hline
    & \textbf{MA-T} & \textbf{MV-T} & \textbf{IV-T} & \textbf{MA-T} & \textbf{MI-T} & \textbf{MV-T} \\
    \hline
    \textbf{Claude 3.5 Sonnet} & 3.8 & 4.0 & 3.1 & 4.1 & 3.1 & 3.2 \\
    \hline
    \textbf{GPT-4o} & 3.6 & 3.8 & 3.1 & 3.9 & 3.1 & 3.2 \\
    \hline
    \textbf{GPT-4oMini} & 3.6 & 3.7 & 3.0 & 3.7 & 3.1 & 3.2 \\
    \hline
    \rowcolor{gray!15} \textbf{MultiMedia-Agent-1} & 3.3 & 3.9 & 3.2 & 3.9 & 3.2 & 3.2 \\
    \hline
    \rowcolor{gray!25} \textbf{MultiMedia-Agent-2} & 3.3 & 3.9 & 3.0 & 4.0 & 3.2 & 3.1 \\
    \hline
    \rowcolor{gray!40} \textbf{MultiMedia-Agent-3} & 3.9 & 3.9 & 3.6 & 4.2 & 3.3 & 3.5 \\
    \hline
\end{tabular}

\caption{Comparisons of \textbf{preference metrics} for \textbf{video generation tasks}. From top to down: \textit{Video Human Alignment; Video Psychological Appealing; Video Aestheic Score; Audio Human Alignment; Audio Psychological Appealing; Audio Video Alignment.}}
\label{tab:comparison_y1_y3}

\end{table*}

\begin{figure*}[h!]
    \centering
    \includegraphics[width=1\linewidth]{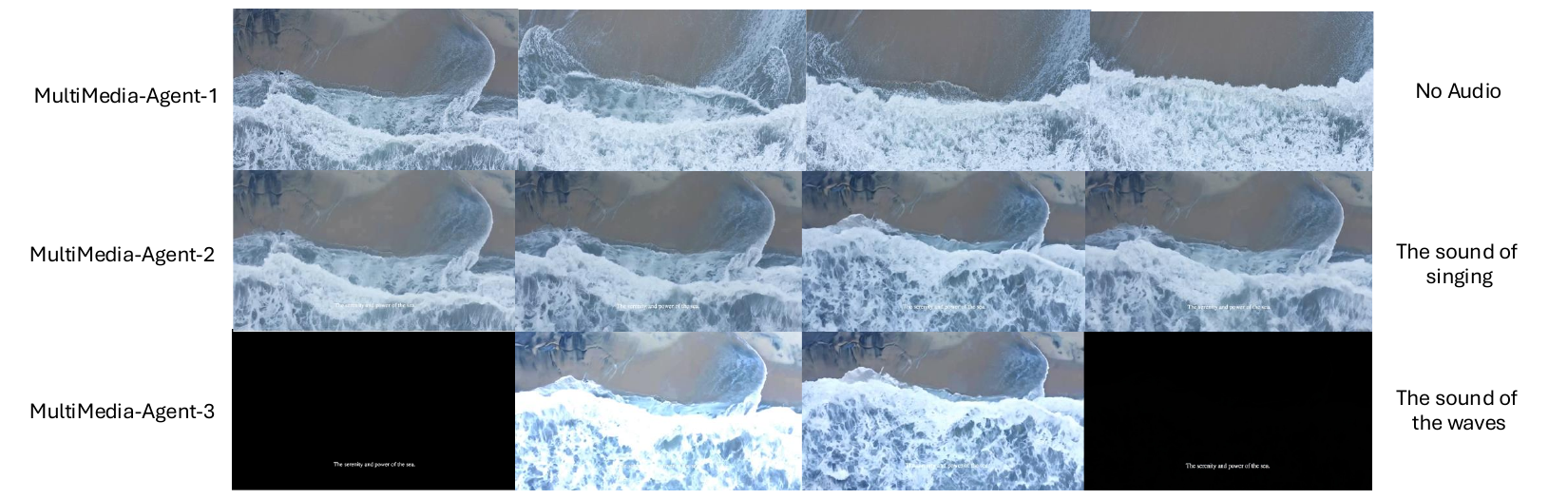}
    \caption{Visualization of the multimedia content created from the plan generated by MultiMedia-Agent. The user query is: use the images and the corresponding video to create a satisfying video.}
    \label{fig:vis}

\end{figure*}

\subsection{Skill Acquisition Theory Can Benefit Tool Agent Training}
We first present a comparison between our MultiMedia-Agent and novel VLMs in terms of success rate in Table~\ref{tab:comparison}. As observed, the agent trained through the Model Associative Stage shows a significant improvement in success rate. However, upon completion of the Model Autonomous Stage, we noticed a decline in success rate. This may be due to the model’s tendency to generate longer plans after the Model Autonomous Stage, and given the model's limited capacity, errors are more likely to occur when generating extended text. This issue is also evident from the comparisons shown in the supplementary materials. This highlights that ensuring the agent outputs a stable plan format is a key challenge for tool agents when dealing with complex scenarios.

We further analyzed the model preference feedback results for content generated by MultiMedia-Agent at different stages compared to GPT-4o. All reported metric results represent the average model feedback scores for successfully executed plans.

Due to space limitations, we only present the results for \textbf{video generation tasks} here, in Table. ~\ref{tab:comparison_y1_y3}. Other results are provided in supplementary material. MultiMedia-Agent-1/2/3 correspond to the agents after each of the three training stages, respectively. As shown in the table, after Stage 1 (Model Cognitive Stage), the MultiMedia-Agent produces fairly average results. Following the second stage of training (Model Associative Stage), the scores dropped, which may be due to the fact that successful plans tend to have fewer steps, leading to weaker alignment.

Moreover, after Stage 3 (Model Autonomous Stage), MultiMedia-Agent showed significant improvements across various metrics. This demonstrates that the rewards from the preference model can effectively optimize the tool agent’s plan generation. Our three-stage training enables the model to effectively learn how to generate complex plans as well as plans aligned with human preferences.

\subsection{Visualization Results}
As seen in Figure \ref{fig:vis}, the plan generated by MultiMedia-Agent-1 lacks corresponding audio and special effects. MultiMedia-Agent-2 added sound effects to the plan, although they did not match the atmosphere of the video. In contrast, MultiMedia-Agent-3 generated content that included subtitles, special effects, and appropriate ocean wave audio. Overall, the visualization results indicate a clear progression in the quality of multimedia generation across different agents. While earlier models struggled with consistency and synchronization, later versions successfully integrated visual, auditory, and textual elements into a cohesive and engaging multimedia output. \textbf{More results are in the supplementary materials}.

\section{Conclusion}

In this paper, we design a multimedia content generation agent system that leverages skill acquisition theory to significantly enhance the capabilities of AIGC technologies in creating complex, multimodal content. By integrating a robust data pipeline, a diverse tool library, and innovative evaluation metrics, our approach not only refines the content generation process but also aligns it more closely with real-world applications. This paves the way for further advancements in automated content creation, promising richer and more effective multimedia outputs.

\subsection{Limitations}

Firstly, for tool selection, we currently use a prompt-based approach. However, considering the vast number of tools available in real-world scenarios, techniques like Retrieval-Augmented Generation (RAG) can be employed to optimize tool selection. Secondly, when solving complex tasks, multi-agent systems are generally more effective than single-agent systems. In our future work, we plan to explore the use of multi-agent systems to tackle complex content generation tasks.

\clearpage 
\bibliographystyle{ACM-Reference-Format}
\bibliography{sample-base}

\clearpage

\end{document}